\documentclass[sigconf, nonacm]{acmart}
\newtheorem{rem}{\bf{Remark}}
\usepackage{makecell}
\usepackage{xcolor}
\definecolor{ForestGreen}{RGB}{34,139,34}

\AtBeginDocument{%
  }

\acmConference[ICAIF '26]{7th ACM International Conference on AI in Finance}{November 14--17, 2025}{Milan, Italy}
\acmISBN{}

\begin{document}

%%
%% The "title" command has an optional parameter,
%% allowing the author to define a "short title" to be used in page headers.
\title{FinSMART: Financial Sentiment Analysis for Algorithmic Trading through Market-Aligned Reinforcement Learning}

\author{Giorgos Iacovides}
\email{giorgos.iacovides20@imperial.ac.uk}
\affiliation{%
  \institution{Imperial College London}
  \city{London}
  \country{UK}
}
\author{Wuyang Zhou}
\email{wuyang.zhou19@imperial.ac.uk}
\affiliation{%
  \institution{Imperial College London}
  \city{London}
  \country{UK}
}

\author{Danilo Mandic}
\email{d.mandic@imperial.ac.uk}
\affiliation{%
  \institution{Imperial College London}
  \city{London}
  \country{UK}
}
\renewcommand{\shortauthors}{Iacovides et al.}

%%
%% The abstract is a short summary of the work to be presented in the
%% article.
\begin{abstract}
%Sentiment extracted from unstructured financial text has become an increasingly important source of information for algorithmic and event-driven trading. 
Recent advances in Generative AI have substantially improved financial sentiment analysis through post-trained financial large language models (LLMs). However, existing approaches remain confined to a market-agnostic, supervised learning paradigm that relies on limited, static and human-annotated datasets, and thus are incapable of adapting to evolving market conditions. To address this limitation, we introduce \emph{FinSMART}, the first market-aligned reinforcement learning framework for financial sentiment analysis, which directly optimizes sentiment signals using realized market outcomes. To deal with the noisy, non-stationary, and mutlifactorial nature of financial markets, FinSMART incorporates a signal extraction pipeline that combines market-aware data filtering with a discrete asymmetric trading reward, enabling stable reinforcement learning from economically meaningful market feedback. Experimental results demonstrate that FinSMART significantly outperforms existing state-of-the-art methods in profitability, risk-adjusted performance, and sentiment signal quality, improving cumulative trading returns by 220\% over the strongest baseline. Uniquely, the FinSMART framework naturally supports market-aware retraining, at any point in time, by replacing costly manual annotation with newly observed financial articles and their realized market outcomes. Such a retraining strategy enables the model to continuously adapt to changing market dynamics, resulting in consistent performance gains over its static counterpart. These findings demonstrate the practical applicability of market-aligned reinforcement learning and highlight its potential as a next-generation paradigm for developing adaptive financial LLMs.
\end{abstract}

\maketitle

\begin{figure}[!t]
\centering
    \includegraphics[width=0.49\textwidth]{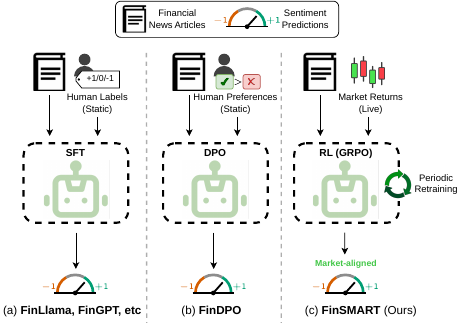}
    %\caption{Financial sentiment analysis approaches. (a) Supervided fine-tuned (SFT)-based and (b) preference-optimized approaches trained on static, labelled datasets. (c) FinSMART `closes the loop' and learns directly from market returns via GRPO, enabling automatic market-aligned predictions and periodic retraining for continual adaptation.
    %}
    \caption{Financial sentiment analysis approaches. (a) Supervised fine-tuning (SFT)-based and (b) preference-optimized approaches rely on static, labelled datasets (c) FinSMART `closes the loop' and learns directly from market returns through GRPO, enabling market-aligned predictions and continual adaptation.}
    \label{finsmart_comp}
\end{figure}

\section{Introduction}
Sentiment analysis aims to quantify market-relevant information embedded within unstructured textual data, such as financial news and earnings reports, and has become an increasingly important component of algorithmic and event-driven trading strategies. This has been enhanced by the rapid advancement of Generative AI (GenAI), particularly Large Language Models (LLMs), which makes it possible to process and interpret vast amounts of unstructured data. Indeed, such data is estimated to account for approximately 80\% of the total data generated within financial markets. Consequently, the ability to effectively leverage these technologies to extract actionable insights from complex narratives and convert them into autonomous trading signals can provide a significant informational advantage, leading to improved price prediction \cite{stock_prices} and the development of effective trading strategies \cite{trad_strats, finllama,findpo}. \par 
Despite conceptual benefits, the diverse, nuanced, and domain-specific nature of financial text presents significant challenges for reliable and robust sentiment extraction. Recently, fine-tuned LLMs \cite{fingpt, zhang2023instructfingpt, finllama, findpo}, specifically adapted to the language of financial markets, have emerged as a promising approach for financial sentiment analysis, owing to their ability to capture the inherent complexities of financial text without requiring the extensive computational resources associated with pre-training. Despite some success, current models remain constrained within a \textit{market-agnostic, supervised learning} framework, as they rely heavily on the availability of limited and static labelled datasets while failing to  incorporate evolving market conditions and real-time dynamics. It is therefore natural to ask:
\begin{itemize}
    \item \textit{Can we develop a financial sentiment analysis framework that extends beyond static labelled training data; can this be achieved by incorporating real-time market dynamics, thereby aligning textual sentiment with subsequent market actions?}
    %item \textit{Can we develop a framework that goes beyond static supervised training and incorporates real-time market dynamics to adapt sentiment representations, thereby bridging the gap between textual sentiment and subsequent price action?}\\
    % \textit{\color{blue}{Can we train a financial sentiment analysis model in a self-supervised way via converting real-time market dynamics into labels (rewards)?}}
\end{itemize}
To this end, we introduce \textit{FinSMART}, the first \underline{Fin}ancial \underline{S}entiment analysis framework based on \underline{M}arket-\underline{A}ligned \underline{R}ein\-force\-ment Learning \underline{T}raining. This allows us to integrate \textit{realized market outcomes}, via idiosyncratic and market returns, directly into the reward function, thus enabling the model to align its extracted sentiment predictions with true market behaviour. \par 
%, rather than relying on static human annotations. \par
\begin{figure*}[!t]
\centering
    \includegraphics[width=1\textwidth]{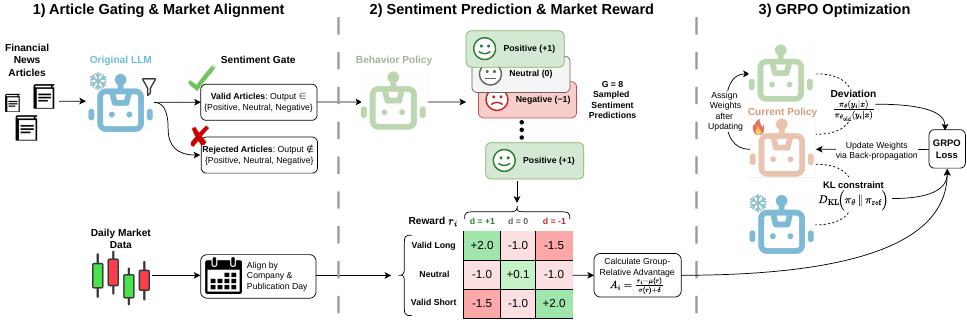}
    \caption{The FinSMART framework for market-aligned reinforcement learning. FinSMART transforms realized market outcomes into a direct supervision signal via a three-stage pipeline: (1) Data Alignment: Financial news articles are filtered and synchronized with publication-day market data to construct clean training pairs; (2) Dual-Filter Reward Evaluation: The behavior policy LLM generates candidate sentiment predictions, which are scored using the proposed dual-filter trading reward to isolate economically meaningful feedback from market noise; and (3) Policy Optimization: Group-relative advantages are optimized via GRPO under a KL divergence constraint against the frozen reference model, thereby enabling sentiment signals to be learned directly from market feedback. 
    } 
    \label{train_grpo}
\end{figure*}
Specifically, a novel dual-objective reward mechanism is designed to overcome the fundamental challenge of extracting meaningful learning signals from financial markets, where returns are inherently noisy, non-stationary, heavy-tailed, and influenced by a number of factors beyond textual sentiment. The proposed reward function integrates a discrete asymmetric trading reward which is designed to explicitly capture the conditions of profitable long and short positions. By structuring the reward signal in this manner, FinSMART reduces sensitivity to noisy market feedback while steering the model towards sentiment predictions that are tightly aligned with subsequent market behavior.
The resulting objective is optimized through a parameter-efficient Group Relative Policy Optimization (GRPO) training framework, enabling efficient reinforcement learning-based adaptation of a pre-trained LLM (specifically Llama-3-8B-Instruct \cite{llama3}) to dynamic market environments. In this way, FinSMART achieves its ultimate goal of enhancing financial sentiment analysis, while also minimizing computational resource requirements. 
The main contributions of this work are therefore:
\begin{itemize}
    \item We introduce FinSMART, the first LLM that directly aligns sentiment predictions with market feedback, rather than relying on static and expensive human labels.
    \item The proposed framework enables, for the first time, a continual adaptation of pre-trained LLMs directly from noisy financial market data. This eliminates the reliance on the scarce, market-agnostic, and costly labeled datasets used by current approaches. 
    \item The FinSMART approach does not require vast computational resources and can operate on a standard GPU. By leveraging a pre-trained LLM and employing parameter-efficient techniques within the GRPO  training, the computational demands typically associated with post-training reinforcement learning are dramatically reduced.
    \item Simulations demonstrate that FinSMART achieves substantial improvements in performance over existing state-of-the-art models across real-world financial metrics. Furthermore, market-aware retraining consistently outperforms its static counterpart, demonstrating the effectiveness of continual adaptation in evolving financial markets.
\end{itemize}
\begin{rem}
Existing financial sentiment analysis frameworks operate in a market-agnostic manner, whereby models are trained on static, labelled datasets, which are limited in scale and unable to fully capture evolving market dynamics. These models are then deployed to generate sentiment predictions without the means to incorporate feedback from subsequent market outcomes. The FinSMART approach establishes market feedback by employing the realized market outcomes as the optimization signal. This makes it possible for sentiment signals to be learned directly from market behaviour, and completely eliminates the dependence on static human-annotated sentiment datasets, as illustrated in Figures \ref{finsmart_comp} and \ref{train_grpo}.
\end{rem}

\section{Related Work}
\textbf{Financial Sentiment Analysis with LLMs.} The application of large language models to financial sentiment analysis began in 2019 with FinBERT \cite{araci2019finbert}, a domain-adapted version of BERT which was fine-tuned on a relatively small corpus of approximately 4,000 labelled financial samples. While FinBERT demonstrated improved performance over general-purpose language models, it also suffers from limitations such as insensitivity to numerical values and reduced robustness to an increase in sentence complexity, due to its relatively small size (110 million parameters). \cite{finbert_complexity}. \par 
More recently, FinGPT \cite{fingpt} and Instruct-FinGPT \cite{zhang2023instructfingpt} introduced a shift towards larger and more general-purpose language models by adapting Llama-7B and Llama-2-13B, respectively, through supervised instruction tuning. Although these models represent an important step towards more capable financial language models, FinGPT was not specifically optimized for financial sentiment analysis. Furthermore, both approaches primarily focus on categorical sentiment prediction (i.e., positive, negative, or neutral), without quantifying sentiment strength, an essential component for downstream applications such as portfolio construction and signal generation. \par 
In contrast, the recent FinLlama \cite{finllama} combines supervised fine-tuning (SFT) with the Llama-2-7B architecture and introduces an additional classification head to generate continuous sentiment scores. While this modification enables sentiment signals to be directly incorporated into portfolio construction pipelines, it also changes the primary function of the model from next token generation to classification. This limits compatibility with advanced post-training techniques that rely on the generative capabilities of LLMs. Furthermore, despite supervised fine-tuning becoming the \textit{de facto} standard for enhancing the performance of pre-trained LLMs on financial sentiment classification tasks, recent studies have highlighted that SFT-based alignment can encourage memorization of training examples and limit generalization to unseen data \cite{rl_sft, findpo}. This is a critical limitation in financial applications, where adaptation to unseen market events is essential  for
robust algorithmic trading strategies. \par 
To address the limitations of SFT-based approaches, Iacovides \textit{et al.} introduced FinDPO \cite{findpo}, which applies direct preference optimization (DPO) during post-training to align large language models with human preferences, thus extending beyond the supervised fine-tuning paradigm. Importantly, FinDPO introduces a logit-to-score conversion mechanism that transforms discrete sentiment predictions into continuous scores, allowing the generated signals to be incorporated directly into portfolio construction. However, FinDPO remains dependent on static labelled datasets, limiting its adaptability to evolving market conditions and preventing direct alignment between sentiment predictions and realized market outcomes. \par 
Motivated by the limitation of existing methods which rely heavily on scarce, expensive, and manually annotated datasets, while remaining largely market-agnostic, we propose FinSMART, the first finance-specific LLM framework for sentiment analysis based on reinforcement learning from market feedback. By framing financial sentiment analysis as a market-aligned reinforcement learning problem, FinSMART derives its learning signal directly from realized market outcomes, thus enabling continual adaptation of the model while reducing dependence on static human-labelled data. Overall, this introduces a new paradigm for financial sentiment analysis, whereby sentiment representations are optimized not only for linguistic consistency but also for their ability to capture economically meaningful market signals.
\section{Preliminaries}
Group Relative Policy Optimization (GRPO) \cite{deepseek-math} is a reinforcement learning-based post-training technique designed to optimize LLMs without the need for a separate value model. Unlike traditional Proximal Policy Optimization (PPO) \cite{ppo}, which relies on a learned critic to estimate state values, GRPO evaluates a group of responses sampled from the policy model and constructs the optimization objective from their relative rewards. \par
Formally, let $\pi_\theta$ denote the policy LLM parameterized by $\theta$, which defines an autoregressive distribution over token sequences (responses) $y$ conditioned on a prompt $x$. Let $R(x,y)\in\mathbb{R}$ be a task-specific reward function, $G$ the number of responses sampled per prompt (the group size), $\pi_{\theta_{\text{old}}}$ the behavior policy used to generate the samples for the current update, and $\pi_{\text{ref}}$ a fixed reference policy, namely the initial pre-trained LLM. \par 
Given a prompt, $x$, drawn from the data distribution, $\mathcal{D}$, the GRPO samples a group of $G$ responses from the behavior policy
\begin{equation}
    \{y_1,y_2,\dots,y_G\} \sim \pi_{\theta_{\text{old}}}(\cdot\mid x),
\end{equation}
whereby each response is assigned a scalar reward, $r_i = R(x,y_i)$. Rather than relying on absolute reward values, the rewards are normalized within the sampled group to obtain relative advantages
\begin{equation}
A_i=\frac{r_i-\mu(r)}{\sigma(r)+\delta},
\end{equation}
where $\mu(r)$ and $\sigma(r)$ designate the mean and standard deviation of the rewards $\{r_1,\dots,r_G\}$ within the group, and $\delta>0$ is a small constant introduced for numerical stability. Each advantage, $A_i$, is assigned to all tokens of its corresponding response, $y_i$.

The policy is optimized using a clipped surrogate objective, regularized by the Kullback--Leibler (KL) divergence from the fixed reference policy $\pi_{\text{ref}}$, such that
\begin{equation}
\begin{aligned}
L_{\text{GRPO}}(\theta)=
-\,\mathbb{E}_{x\sim\mathcal{D},\,\{y_i\}\sim\pi_{\theta_{\text{old}}}}
\bigg\{
\frac{1}{G}\sum_{i=1}^{G}
\min\Big\{
\tfrac{\pi_\theta(y_i\mid x)}{\pi_{\theta_{\text{old}}}(y_i\mid x)}\,A_i,\; \\
\text{clip}\Big(
\tfrac{\pi_\theta(y_i\mid x)}{\pi_{\theta_{\text{old}}}(y_i\mid x)},
1-\epsilon,1+\epsilon
\Big)A_i
\Big\}
-\beta\,D_{\text{KL}}\!\big(\pi_\theta \,\|\, \pi_{\text{ref}}\big)
\bigg\},
\label{eq:grpo_objective}
\end{aligned}
\end{equation}
where the policy ratio $\frac{\pi_\theta(y_i\mid x)}{\pi_{\theta_{\text{old}}}(y_i\mid x)}$ measures the deviation of the current policy, $\pi_\theta$, from the behavior policy, $\pi_{\theta_{\text{old}}}$, the clipping range, $\epsilon$, bounds the per-update change of $\pi_\theta$ relative to $\pi_{\theta_{\text{old}}}$, and $\beta$ weighs the KL penalty that keeps the updated policy close to the fixed reference policy, $\pi_{\text{ref}}$. The typical GRPO training procedure is as follows:

\begin{enumerate}
    \item For each prompt, $x$, sample a group of $G$ candidate responses from the behavior policy, $\pi_{\theta_{\text{old}}}$.
    \item Score each response with the reward function and normalize the rewards within the group to obtain the relative advantages, $A_i$.
    \item Update the current policy, $\pi_\theta$, to increase the likelihood of higher-advantage responses and decrease that of lower-advantage ones, subject to the clipping constraint on the policy ratio and the KL constraint toward $\pi_{\text{ref}}$.
\end{enumerate}

Intuitively, GRPO avoids the need for absolute reward calibration by learning from the relative quality of generated outputs within each sampled group. This property makes it particularly suitable for settings where the reward signal is noisy or difficult to model explicitly, as is typically the case in financial markets. It is important to note that, although GRPO has recently gained attention for improving reasoning capabilities in LLMs, its underlying optimization mechanism is not limited to reasoning tasks. Moreover, recent studies suggest that reasoning, whether prompt-induced or built-in, does not improve financial sentiment accuracy \cite{reasoning_fsa}. Therefore, in this work, we leverage GRPO primarily as an efficient reinforcement learning optimization framework, in which multiple sampled generations enable exploration over diverse sentiment predictions and allow the model to optimize directly against a market-based reward signal.
\section{Training Framework of FinSMART} \label{training_framework}
Our work seeks to leverage the contextual understanding of general-purpose LLMs and adapt them to financial sentiment analysis through reinforcement learning from market feedback. Rather than learning from static, manually annotated, and often simplified labelled examples, FinSMART is trained on real-world financial articles and optimized using GRPO, whereby realized market returns provide the learning signal. For rigor, our model is evaluated over a set of benchmarks that closely align with real-world portfolio construction — the ultimate goal of sentiment analysis. The complete training framework of FinSMART is illustrated in Figure \ref{train_grpo} and outlined below. 
\subsection{Data Pipeline and Signal Extraction} \label{data_pipeline}
% \subsubsection{Data Pipeline and Signal Extraction}
Given the highly noisy, non-stationary, and heavy-tailed nature of financial returns, together with the fact that market movements are driven by numerous factors beyond textual sentiment, extracting a reliable learning signal from realized returns is inherently challenging. The central premise of FinSMART is therefore that reinforcement learning from market feedback is only effective when the policy is trained on articles that (i) contain a \emph{decodable} sentiment signal—that is, articles from which the underlying LLM can confidently infer a valid sentiment prediction—and (ii) can be reliably aligned with a \emph{realized} market outcome. Consequently, data construction is not merely a pre-processing step, but an integral component of the proposed signal extraction framework. \newline 
\noindent\textbf{Data Collection.} Financial news articles were collected from The Motley Fool (TMF) and MarketWatch over the period from February 2015 to June 2021. These sources were selected due to their reputation, low reporting bias, and extensive coverage of publicly traded companies. The corresponding daily stock market data were obtained from Yahoo Finance for all companies in the S\&P 500, which constitutes our investable universe. This resulted in 1,672 trading days of return data for each company. \newline 
\noindent\textbf{Named Entity Recognition (NER).} A prerequisite for learning from market feedback is ensuring that each news article is associated with the correct traded asset. Since financial articles frequently discuss multiple companies, naively linking an article to a single stock can introduce substantial misalignment into the reward signal. To address this, we applied Named Entity Recognition (NER) to identify the primary organizational entities referenced within each article. Specifically, we employed the BERT-base-NER model \cite{bert_base_ner}, which recognizes four entity types (organization, person, location, and miscellaneous) and produces a confidence score for each detected entity. Articles were retained only when the confidence score of the target company exceeded 98\%; otherwise, they were discarded. This filtering step substantially reduced the risk of assigning market returns to unrelated articles, thereby improving the reliability of the market-derived reward used during our RL training. \newline 
\noindent\textbf{Sentiment Gating.} A large fraction of financial text does not contain a clear sentiment opinion. Rather than relying on manually designed heuristics to remove such articles, we use the reference policy, $\pi_{\text{ref}}$, as a \emph{sentiment gate}. In particular, each entity-linked article is passed through the pre-trained LLM under a constrained classification prompt that restricts the output space to a single label from $\{\texttt{Positive}, \texttt{Negative}, \texttt{Neutral}\}$. Articles for which the model does not assign one of the three sentiment labels as the most probable next token are discarded. Intuitively, these cases correspond to inputs for which the reference model does not produce a confident sentiment signal, and retaining them would introduce noise into the reward signal.
\subsection{Dual-Filter Trading Reward}
The effectiveness of reinforcement learning depends critically on the quality of the reward signal. In financial markets, however, realized returns constitute an inherently challenging supervision signal. Consequently, naively rewarding a model according to subsequent returns would expose the policy to an extremely low signal-to-noise ratio, leading to unstable optimization and poor generalization. \par 
To address this challenge, we design a market-aligned reward function whose objective is twofold: (i) to extract economically meaningful learning signals from noisy market outcomes, and (ii) to provide a stable optimization landscape that avoids degenerate solutions (mode collapse) during training. \par 
Given an article, $x$, the policy generates a sentiment prediction $d \in \{-1,0,+1\},$
corresponding to a negative, neutral or positive market outlook. Let $\alpha$ denote the idiosyncratic return of the stock on \emph{publication} day, computed as the stock's realized return minus the return of the S\&P 500 index, proxied by the SPY ETF, i.e., $\alpha = r_{\text{stock}} - r_{\text{SPY}}$, and let $r$ denote the stock's realized raw return. To formalize the symmetric reward structure, we first map the market outcomes to a ground-truth directional label, $y \in \{-1, 0, +1\}$, as
\[
y = \begin{cases} 
+1, & \text{if } \alpha > \tau \text{ and } r > 0, \\ 
-1, & \text{if } \alpha < -\tau \text{ and } r < 0, \\ 
0, & \text{otherwise}, 
\end{cases}
\]
where $\tau$ denotes the minimum alpha threshold required to regard the market reaction 
as economically meaningful. Throughout this work we set $\tau=0.5\%$. The reward assigned to each sampled completion is then defined by grouping the behavioral 
alignment between the predicted sentiment $d$ and the true market state $y$, in the form
\[
R_{\text{trade}}(d, y) = \begin{cases} 
+2.0, & \text{if } d = y \text{ and } y \neq 0 \quad (\text{Correct Direction}), \\[1.5mm] 
+0.1, & \text{if } d = y = 0 \quad (\text{Correct Neutral}), \\[1.5mm] 
-1.5, & \text{if } d = -y \text{ and } y \neq 0 \quad (\text{Opposite Direction}), \\[1.5mm] 
-1.0, & \text{o/w} \quad (\text{Missed Trade or False L/S Signal}). 
\end{cases}
\]
%The reward assigned to each sampled completion is defined as
% \[
% R_{\text{trade}}(d,\alpha,r)=
% \begin{cases}
% +2, & d=+1,\; \alpha>\tau,\; r>0,\\[2mm]
% +2, & d=-1,\; \alpha<-\tau,\; r<0,\\[2mm]
% -1, & d=+1,\; \alpha<-\tau,\; r<0,\\[2mm]
% -1, & d=-1,\; \alpha>\tau,\; r>0,\\[2mm]
% 0, & \text{otherwise},
% \end{cases}
% \]
% where $\tau$ denotes the minimum alpha threshold required to regard the market reaction as economically meaningful. Throughout this work we set $\tau=0.5\%$. \par 
Unlike conventional classification rewards, the proposed objective requires a sentiment prediction to satisfy two independent market conditions to receive positive reinforcement. First, the predicted direction must agree with the realized return, ensuring the implied trading position is profitable. Second, the corresponding alpha return must exceed a minimum threshold, ensuring the observed price movement is attributable to stock-specific information rather than broad market fluctuations. This dual-filter design reduces the influence of spurious price movements and extracts a more informative learning signal from inherently noisy financial data. \par 
Furthermore, the reward is intentionally asymmetric, so that correct predictions receive a larger positive reward than the penalty assigned to incorrect predictions. This encourages exploration during policy optimization while avoiding overly conservative behaviour, where the model collapses towards predominantly neutral predictions in response to uncertain market feedback. \par 
It is important to note that, during training, rewards are computed using the \emph{same-day} raw and alpha returns associated with each article. While this uses contemporaneous market information to construct the reward, our objective is to maximize the quality of the supervisory signal rather than simulate a trading strategy during training. Empirical analysis demonstrates that publication-day returns exhibit substantially stronger alignment with article sentiment than one-day-ahead returns across all data sources.
For example, on TMF the average publication-day alpha differs by approximately $5.0\%$ between articles classified as positive and negative by the pre-trained reference model, whereas this separation falls to just $0.3\%$ when returns are shifted by one trading day. A similar reduction is observed for MarketWatch, where the corresponding spread decreases from $2.3\%$ to $0.3\%$. Likewise, the Pearson correlation between reference sentiment and alpha returns decreases from $0.41$ to $0.03$ on TMF and from $0.37$ to $0.03$ on MarketWatch when next-day returns are used. These results indicate that publication-day returns provide a substantially richer and less noisy signal for the GRPO training. To ensure a fair evaluation, all trading experiments use next-day returns, to simulate realistic  execution and avoid look-ahead bias.

\begin{rem}
    Given that the reward is deliberately designed to maximise the signal-to-noise ratio of market feedback, same-day returns are used exclusively during training due to their substantially stronger alignment with article sentiment compared to one-day-ahead returns. All reported trading results are evaluated using next-day returns to eliminate look-ahead bias.
\end{rem}
\subsection{GRPO Training Configuration}

The proposed training configuration is designed to ensure stable reinforcement learning despite the noisy and stochastic nature of market-derived rewards. Since rewards are computed from realized market outcomes, their magnitude alone provides a weak learning signal. Instead, GRPO employs group-relative normalization, allowing the policy to learn from the \emph{relative} ranking of candidate sentiment predictions rather than from their absolute rewards. Consequently, optimization remains critic-free while providing a substantially denser and more stable learning signal.

To encourage exploration, multiple sentiment completions are sampled for each training article. Specifically, we sample $G=8$ completions from the current policy, which are used to compute the GRPO objective in Eq.~\eqref{eq:grpo_objective} together with a KL regularization term of $\beta=0.1$. Sampling multiple responses enables the policy to explore alternative sentiment predictions for the same market event, while the proposed asymmetric dual-filter trading reward encourages convergence towards economically meaningful trading decisions rather than premature collapse to a single sentiment label. \par
To minimize computational requirements, we employed Low-Rank Adaptation (LoRA) \cite{lora} during GRPO training, with rank $r=16$, scaling factor $\alpha=32$, and a dropout rate of $0.05$. This reduced the number of trainable parameters to just $13.6$M (approximately $0.17\%$ of the  base model parameters), allowing the entire RL pipeline to be performed on a single NVIDIA A6000 (48 GB) GPU.
\par 
For all experiments, articles published up to 31 December 2018 were used for training, while articles published between January 2019 and June 2021 were reserved exclusively for out-of-sample inference and backtesting. This chronological split serves two purposes. First, it provides a sufficiently long evaluation period, including the COVID-19 market crash, enabling the robustness of FinSMART to be assessed under both normal and highly volatile market conditions. Second, it preserves approximately 30,000 training articles, ensuring a training corpus of comparable size to prior financial sentiment models \cite{finllama, findpo}, thereby facilitating a fair comparison. Within the training set, 5\% of the training data was held out as a validation set, which was used for model selection, with evaluation performed every 500 training steps. Under this configuration, the training run was completed in 8 hours.
% \subsubsection{Source-Specific Adapters}
% Separate LoRA adapters are trained for TMF and MarketWatch to account for the substantial stylistic differences between the two sources. TMF primarily publishes long-form editorial analyses of individual companies, whereas MarketWatch produces shorter, news-driven articles covering a broader set of firms. We hypothesize that training source-specific adapters allows each model to specialize to the linguistic characteristics of its respective corpus while preserving the shared knowledge encoded within the underlying LLM.

\section{Sentiment-Driven Portfolio Construction Framework} \label{port_construction}
\begin{figure}[!h]
\centering\includegraphics[width=0.48\textwidth]{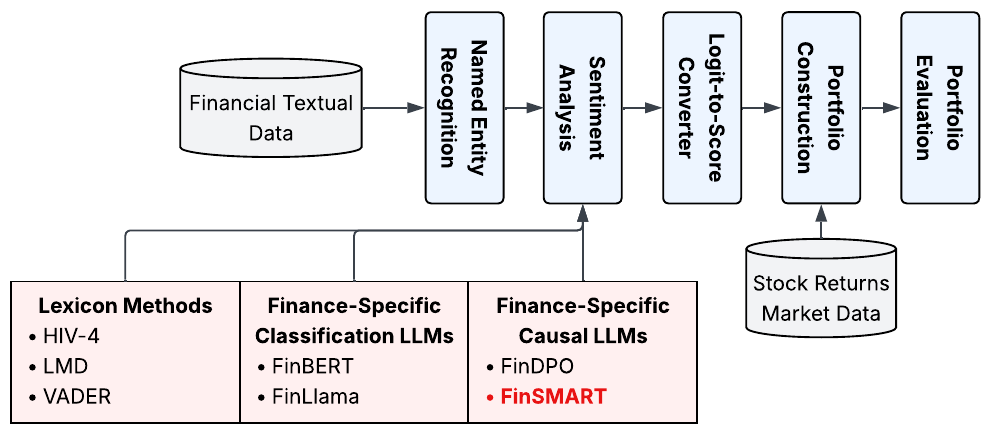}
    \caption{Proposed framework for our sentiment-driven portfolio construction. Financial articles ($N \approx 325,000$) from the out-of-sample period (January 2019--June 2021) are used to avoid look-ahead bias. Named Entity Recognition (NER) is performed as described in Section \ref{data_pipeline}. The `logit-to-score' converter is only required for finance-specific causal LLMs.}
    \label{fig:framework}
\end{figure}

Once our FinSMART model was established, the extracted sentiment signals were subsequently integrated into a portfolio construction framework to assess their economic value. Since FinSMART is built upon a causal LLM, sentiment predictions are converted into continuous sentiment scores using the logit-to-score converter proposed in \cite{findpo}. This enabled direct comparison with existing financial sentiment analysis methods within a realistic portfolio construction setting, where performance was evaluated using finance-specific, economically meaningful metrics. The overall evaluation framework is illustrated in Figure~\ref{fig:framework}.
\begin{figure*}[!t]
\centering
    \includegraphics[width=1.0\textwidth]{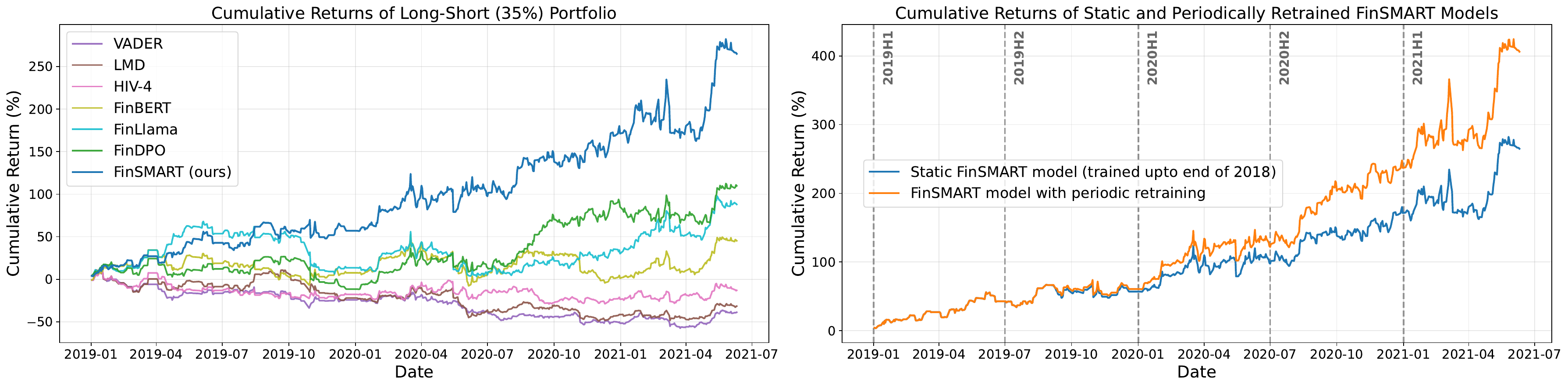}
    \caption{Cumulative returns of sentiment-based long-short portfolios and the impact of market-aligned retraining. Left: Out-of-sample cumulative returns of the 35\% long-short portfolios constructed. Right: Comparison between the static FinSMART model (trained using data up to December 2018) and the periodically retrained FinSMART model. As retraining occurs every six months, both models are identical during the first six months (2019H1) before the first model update.}
    \label{fig:results}
\end{figure*} \newline 
\noindent \textbf{Sentiment Analysis.} We compared FinSMART against six widely used financial sentiment analysis methods, spanning lexicon-based approaches, supervised fine-tuned (SFT) LLMs, and preference-optimized LLMs. In particular, we included LMD \cite{lmd}, HIV-4 \cite{hiv4}, and VADER \cite{vader} as representative lexicon-based methods, together with FinBERT \cite{araci2019finbert}, FinLlama \cite{finllama}, and FinDPO \cite{findpo}. Notably, FinDPO represents the current state-of-the art model for financial sentiment analysis and was the first causal LLM designed specifically for portfolio construction. The LMD and HIV-4 lexicons were implemented using the \texttt{pysentiment2} library, VADER was implemented using \texttt{NLTK}, and all LLM-based baselines were obtained from HuggingFace and evaluated using the \texttt{Transformers} library. \par 
Given that FinSMART is built upon a causal LLM, its outputs are naturally limited to discrete sentiment labels (i.e., positive, negative, or neutral). To obtain continuous sentiment scores suitable for portfolio construction, we employed the logit-to-score converter proposed in FinDPO, which leverages the internal output logits of the model to quantify the strength of each sentiment prediction without requiring any architectural modifications. The considered methods were evaluated on all articles drawn from an out-of-sample corpus of size $N\approx 325,000$.  In cases where multiple articles
were published on the same day for a given company, their sentiment scores were aggregated by computing the daily average, \begin{equation} S_t=\frac{1}{N_t}\sum_{i=1}^{N_t}S_{it}, \end{equation} where $S_t$ denotes the aggregate sentiment score for day $t$, $N_t$ is the number of articles published for the company on that day, and $S_{it}$ is the sentiment score assigned to the $i$-th article. 
\noindent \textbf{Portfolio Construction.} Once the sentiment score for each method was defined for every company, a daily long--short portfolio was constructed. We used the sentiment strength as a parameter to determine which companies should be in a long or a short position. The portfolio construction procedure is summarized below:
\begin{itemize}
\item \textit{Define the Investable Universe.} Although the benchmark universe is based on S\&P 500 constituents, the investable universe is constructed in a point-in-time and dynamic manner. On each trading day, only companies with point-in-time financial news and corresponding market data were included. This daily reconstitution naturally accounts for entries, delistings, mergers, and acquisitions throughout the sample period, thereby mitigating survivorship bias.
\item \textit{Define the long and short positions.} The sentiment signals produced by each of the seven methods were used to construct seven independent portfolios. On each trading day, companies were ranked according to their sentiment score. Companies that did not have sentiment data on a particular day were omitted from the ranking. As the daily sentiment score for each company ranges between -1 and 1, companies with the highest positive sentiment were placed in a long position, while those with the strongest negative sentiment were placed in a short position. 
\item \textit{Portfolio allocation.} Following \cite{finllama,findpo}, an equally weighted 35\% long--short portfolio was employed, with the top 35\% of ranked companies allocated to long positions and the bottom 35\% allocated to short positions. This portfolio construction methodology is consistent with the long--short strategies commonly employed by equity hedge funds \cite{hedge_funds_strat}.
\item \textit{Determine daily returns.} To ensure a realistic trading setting and eliminate look-ahead bias, sentiment generated from articles published on trading day $t$ was used to construct the portfolio at the market open of the next trading day $t+1$. Positions were then held for one trading day and closed at the subsequent market open, implying that all returns were computed on a next-day open-to-open basis. The average daily return of the long portfolio was computed as
\begin{equation} 
r_{\mathrm{Long},t+1} = \frac{1}{N_{\mathrm{Long}}} \sum_{i=1}^{N_{\mathrm{Long}}} r_{i,t+1}. \end{equation} 
Similarly, the average daily return of the short portfolio was computed as 
\begin{equation} r_{\mathrm{Short},t+1} = \frac{1}{N_{\mathrm{Short}}} \sum_{i=1}^{N_{\mathrm{Short}}} r_{i,t+1}. 
\end{equation}
For each particular day, the number of companies that were held in either a long position ($N_{Long}$) or a short position ($N_{Short}$) were equal. Consequently, the total portfolio return on a particular day is the difference between the daily long return and the daily short return, and is given by
\begin{equation} 
r_{t+1} = r_{\mathrm{Long},t+1} - r_{\mathrm{Short},t+1}. \end{equation}
\end{itemize}

\noindent \textbf{Portfolio Evaluation.} The performance of the portfolio constructed using the proposed model was assessed against the portfolios constructed using the other sentiment methods. We considered:

\begin{itemize}
    \item \textit{Profitability}, measured via cumulative returns, $r_{\text{cum}}$, and annualized returns, $R_p$, defined as 
    \begin{equation}
    r_{\text{cum}} = \sum_{i=1}^{N} r_{\text{daily}}(i),
    \end{equation}
    \begin{equation}
    R_p = 252 \cdot \frac{1}{N} \sum_{i=1}^{N} r_{\text{log}}(i).
    \end{equation}

    \item \textit{Risk-adjusted performance}, evaluated using the annualized Sharpe ratio $S_a$, Sortino ratio $S_o$, and Calmar ratio $C_r$, defined as
    \begin{equation}
    S_a = \frac{R_p - R_f}{\sigma_p},
    \end{equation}
    \begin{equation}
    S_o = \frac{\bar{r}_{\text{daily}} - R_f}{\sigma_d} \cdot \sqrt{252},
    \end{equation}
    \begin{equation}
    C_r = \frac{(1 + \bar{r}_{\text{simple}})^{252} - 1}{\text{MDD}}.
    \end{equation}
\end{itemize}
Here, \( N \) is the total number of trading days, \( r_{\text{log}}(i) \) is the daily logarithmic return, and \( \bar{r}_{\text{simple}} \) is the average daily simple return. 
The symbol \( R_f \) denotes the annualized risk-free rate of return, \( \sigma_p \) is the annualized volatility, \( \sigma_d \) is the downside deviation of daily returns, and MDD is the maximum drawdown. The constant 252 corresponds to the number of business days in a calendar year. In line with the literature, and given the persistently low yields during the sample period \cite{low_yield}, we set $R_f=0$.
\section{Experimental Results}
The performance of the seven sentiment-based portfolios, constructed as described in Section~\ref{port_construction}, is presented in the left panel of Figure~\ref{fig:results} and in Table~\ref{tab:performance_0bps}.
\subsection{Evaluation via Real-World Financial Metrics} \label{eval_metrics}
Our proposed method, FinSMART, consistently outperformed all competing sentiment analysis methods across every evaluation metric. Notably, despite using the same base model as FinDPO, the existing state-of-the-art model in financial sentiment analysis, FinSMART achieved a 141\% improvement in cumulative returns (264.9\% vs.\ 109.8\%) and more than doubled the annualized returns (91.5\% vs.\ 45.0\%). 
%Beyond profitability, the proposed approach also delivers superior risk-adjusted performance. Specifically, it improved the Sharpe ratio from 1.12 to 1.97, indicating a substantially more favourable return-to-volatility trade-off. Moreover, it attained significantly higher Sortino (2.40) and Calmar (4.23) ratios, highlighting superior downside risk control and greater resilience to drawdowns while maintaining substantially higher returns. \par 
Beyond profitability, the proposed approach also delivered superior risk-adjusted performance, improving the Sharpe ratio from 1.12 to 1.97, thus indicating a more favourable return-to-volatility trade-off. Furthermore, it attained significantly higher Sortino (2.40) and Calmar (4.23) ratios, demonstrating improved downside risk control and greater resilience to drawdowns.
To further evaluate whether the improvement in portfolio performance is driven by a more informative sentiment signal, we assessed the cross-sectional predictive ability of each sentiment model using the Rank Information Coefficient (RankIC). Table~\ref{tab:performance_0bps} shows that FinSMART achieved the highest RankIC, with a 15\% improvement over FinDPO (0.061 vs.\ 0.053). This indicates that the sentiment signals learned through market feedback exhibit stronger alignment with subsequent stock returns. 
\par 
 % Simultaneously, we evaluate the quality of the underlying sentiment signal using the Rank Information Coefficient (RankIC). FinSMART achieves a 15\% improvement over FinDPO (0.062 vs.\ 0.054), indicating a stronger cross-sectional alignment between predicted sentiment and subsequent market returns. \par 
Overall, these results demonstrate that FinSMART not only generates substantially higher returns than existing sentiment-based approaches, but also produces more stable and risk-efficient portfolios driven by a higher-quality sentiment signal. 

\newcommand{\posval}[1]{\textcolor{ForestGreen}{#1}}
\newcommand{\negval}[1]{\textcolor{red}{#1}}
\begin{table}[h!]
\centering
\caption{Performance comparison of the seven sentiment-based portfolios and their underlying predictive signal quality. Portfolio metrics are evaluated using out-of-sample trading returns, while RankIC measures the cross-sectional alignment between sentiment scores and next-day alpha returns. For each metric, the best performance is shown in bold, and the second-best is underlined.}
\label{tab:performance_0bps}
\resizebox{\linewidth}{!}{%
\begin{tabular}{lcccccc}
\specialrule{.1em}{0.2em}{0.3em}
\textbf{Method} & 
\makecell{\textbf{Cumulative}\\\textbf{Return (\%)}} & 
\makecell{\textbf{Annualized}\\\textbf{Return (\%)}} & 
\textbf{Sharpe} & 
\textbf{Sortino} & 
\textbf{Calmar} &
\textbf{RankIC} \\
\specialrule{.1em}{0.2em}{0.3em}
\textbf{S\&P 500}  & \posval{69.3} & \posval{26.9} & \posval{1.09} & \posval{1.18} & \posval{0.79} & -- \\
\specialrule{.1em}{0.2em}{0.3em}
HIV-4        & \negval{$-$13.0} & \negval{$-$6.8}  & \negval{$-$0.03} & \negval{$-$0.07} & \negval{$-$0.18} & +0.006 \\
VADER        & \negval{$-$38.8} & \negval{$-$21.9} & \negval{$-$0.45} & \negval{$-$0.57} & \negval{$-$0.35} & +0.011 \\
LMD          & \negval{$-$31.6} & \negval{$-$17.3} & \negval{$-$0.28} & \negval{$-$0.37} & \negval{$-$0.32} & +0.029 \\
FinBERT      & \posval{45.2} & \posval{20.6} & \posval{0.67} & \posval{0.86} & \posval{0.52} & +0.043 \\
FinLlama     & \posval{87.9} & \posval{37.3} & \posval{0.98} & \posval{1.24} & \posval{0.97} & +0.048 \\
FinDPO       & \underline{\posval{109.8}} & \underline{\posval{45.0}} & \underline{\posval{1.12}} & \underline{\posval{1.44}} & \underline{\posval{1.48}} & \underline{+0.053} \\
\textbf{FinSMART (Ours)} & 
\textbf{\posval{264.9}} & 
\textbf{\posval{91.5}} & 
\textbf{\posval{1.97}} & 
\textbf{\posval{2.40}} & 
\textbf{\posval{4.23}} & 
\textbf{+0.061} \\
\specialrule{.1em}{0.2em}{0.3em}
\end{tabular}
}
\end{table}

\subsection{Periodic Market-Aligned Retraining}
Section~\ref{eval_metrics} demonstrates that directly incorporating market feedback into the reward function and aligning the model with realized market outcomes leads to substantial improvements in profitability, risk-adjusted performance, and sentiment signal quality, compared with existing approaches trained on static, human-annotated datasets. Since these datasets are inherently market-agnostic and fixed at the time of annotation, they are unable to adapt to the continuously evolving nature of financial markets.\par 
A key distinguishing feature of the proposed FinSMART framework is its ability to naturally support market-aware retraining, unlike existing financial sentiment analysis methods that rely on static labeled datasets. Rather than requiring costly manual annotation, newly published financial articles can be automatically paired with their subsequently realized market returns to generate fresh training data, enabling the model to be periodically re-aligned with evolving market conditions. To demonstrate this capability, we implemented an expanding-window retraining strategy, in which the model was retrained every six months using all data accumulated up to that point.  The six-month update interval provides a practical balance between accumulating a sufficiently large and diverse set of newly observed articles for effective retraining and ensuring that the model remains responsive to evolving market dynamics. \par 
The first retraining was performed using all articles up to the first half of 2019, after which the model was subsequently updated at six-month intervals. At each stage, inference was conducted using the trained model over the following six-month period, and the newly observed articles were then incorporated into the training set in an expanding-window fashion. Given that the corpus of financial articles extends up to June 2021, a total of four model iterations were trained. Importantly, all training procedures were identical across iterations and followed the same setup as the static model described in Section~\ref{training_framework}. The only difference between iterations was the availability of additional data induced by the expanding-window retraining schedule. \par 
The right panel of Figure~\ref{fig:results} compares the cumulative returns of the static model (trained using financial articles up to the end of 2018) with those of the periodically retrained FinSMART model. The retrained model consistently outperforms the static benchmark throughout the evaluation period, increasing cumulative returns from 265\% to 406\%. Furthermore, as shown in Table~\ref{tab:finsmart_static_vs_retrain}, all risk-adjusted performance metrics, together with RankIC, improved consistently. These results demonstrate that periodic market-aligned retraining not only increases portfolio performance, but also continually re-aligns the underlying sentiment model with evolving market dynamics, leading to progressively stronger and more informative trading signals. \par 
\begin{table}[h!]
\centering
\caption{Performance comparison of the static and periodically retrained FinSMART models. For each metric, best performance is shown in bold; second-best is underlined.}
\label{tab:finsmart_static_vs_retrain}
\resizebox{\linewidth}{!}{%
\begin{tabular}{lcccccc}
\specialrule{.1em}{0.2em}{0.3em}
\textbf{Method} &
\makecell{\textbf{Cumulative}\\\textbf{Return (\%)}} &
\makecell{\textbf{Annualized}\\\textbf{Return (\%)}} &
\textbf{Sharpe} &
\textbf{Sortino} &
\textbf{Calmar} &
\textbf{RankIC} \\
\specialrule{.1em}{0.2em}{0.3em}
FinSMART (Static) &
\underline{\posval{264.9}} &
\underline{\posval{91.5}} &
\underline{\posval{1.97}} &
\underline{\posval{2.40}} &
\underline{\posval{4.23}} &
\underline{+0.061} \\
FinSMART (Retrained) &
\textbf{\posval{406.2}} &
\textbf{\posval{125.7}} &
\textbf{\posval{2.41}} &
\textbf{\posval{2.96}} &
\textbf{\posval{5.65}} &
\textbf{+0.065}\\
\specialrule{.1em}{0.2em}{0.3em}
\end{tabular}
}
\end{table}
Interestingly, we observe a moderately strong positive correlation between the number of articles added in each six-month period and the performance gain of the retrained model relative to the static baseline ($r = 0.72$), suggesting that richer incoming information enhances the benefit of periodic retraining.

\section{Conclusion}
We have introduced FinSMART, a market-aligned reinforcement learning framework for financial sentiment analysis. Unlike conventional approaches that rely on static, market-agnostic, and labelled datasets, FinSMART incorporates realized market outcomes into the reinforcement learning objective, enabling sentiment signals to be optimized according to their subsequent economic impact. To address the noisy and non-stationary nature of financial markets, our framework has introduced a signal extraction pipeline that combines market-aware data filtering with a discrete asymmetric trading reward, thus allowing reinforcement learning to optimize against economically meaningful reward signals while maintaining stable training dynamics. By incorporating market feedback directly into the training process, FinSMART has significantly outperformed the existing state-of-the-art model, FinDPO, in profitability, risk-adjusted performance, and RankIC, demonstrating that optimizing language models with realized market outcomes can produce more economically informative sentiment signals and more profitable, risk-efficient trading strategies. \par 
Uniquely, we have demonstrated that FinSMART naturally supports market-aligned retraining. By replacing  static, labelled datasets with newly observed financial articles and their realized market outcomes, the framework can continuously incorporate evolving market feedback and adapt to changing market conditions. Experimental results have highlighted that this retraining strategy consistently outperforms its static counterparts, confirming the practical applicability of FinSMART in real-world financial environments. We believe this paradigm opens new avenues for developing adaptive financial LLMs that continually learn from market behaviour, rather than relying solely on laborious human annotations.

\bibliographystyle{ACM-Reference-Format}
\bibliography{references}

%%
%% If your work has an appendix, this is the place to put it.
%\appendix

% \section{Research Methods}

% \subsection{Part One}

% Lorem ipsum dolor sit amet, consectetur adipiscing elit. Morbi
% malesuada, quam in pulvinar varius, metus nunc fermentum urna, id
% sollicitudin purus odio sit amet enim. Aliquam ullamcorper eu ipsum
% vel mollis. Curabitur quis dictum nisl. Phasellus vel semper risus, et
% lacinia dolor. Integer ultricies commodo sem nec semper.

% \subsection{Part Two}

% Etiam commodo feugiat nisl pulvinar pellentesque. Etiam auctor sodales
% ligula, non varius nibh pulvinar semper. Suspendisse nec lectus non
% ipsum convallis congue hendrerit vitae sapien. Donec at laoreet
% eros. Vivamus non purus placerat, scelerisque diam eu, cursus
% ante. Etiam aliquam tortor auctor efficitur mattis.

% \section{Online Resources}

% Nam id fermentum dui. Suspendisse sagittis tortor a nulla mollis, in
% pulvinar ex pretium. Sed interdum orci quis metus euismod, et sagittis
% enim maximus. Vestibulum gravida massa ut felis suscipit
% congue. Quisque mattis elit a risus ultrices commodo venenatis eget
% dui. Etiam sagittis eleifend elementum.

% Nam interdum magna at lectus dignissim, ac dignissim lorem
% rhoncus. Maecenas eu arcu ac neque placerat aliquam. Nunc pulvinar
% massa et mattis lacinia.

\end{document}